\documentclass{article}

\usepackage{PRIMEarxiv}

\usepackage[utf8]{inputenc} 
\usepackage[T1]{fontenc}    
\usepackage{hyperref}       
\usepackage{url}            
\usepackage{booktabs}       
\usepackage{amsfonts}       
\usepackage{nicefrac}       
\usepackage{microtype}      
\usepackage{lipsum}
\usepackage{fancyhdr}       
\usepackage{graphicx}       
\graphicspath{{media/}}     
\usepackage{subcaption}
\usepackage{amsmath}

\title{Unveiling Redundancy in Diffusion Transformers (DiTs): A Systematic Study
}

\author{
  Xibo Sun \\
  Tencent \\
  \texttt{xibosun@tencent.com} \\
   \And
  Jiarui Fang \\
  Tencent \\
  \texttt{jiaruifang@tencent.com} \\
   \And
  Aoyu Li \\
  Tencent \\
  \texttt{aoyuli@tencent.com} \\
   \And
  Jinzhe Pan \\
  Tencent \\
  \texttt{eigenpan@tencent.com} \\
}

\begin{document}
\maketitle

\begin{abstract}
The increased model capacity of Diffusion Transformers (DiTs) and the demand for generating higher resolutions of images and videos have led to a significant rise in inference latency, impacting real-time performance adversely. While prior research has highlighted the presence of high similarity in activation values between adjacent diffusion steps (referred to as redundancy) and proposed various caching mechanisms to mitigate computational overhead, the exploration of redundancy in existing literature remains limited, with findings often not generalizable across different DiT models.
This study aims to address this gap by conducting a comprehensive investigation into redundancy across a broad spectrum of mainstream DiT models. Our experimental analysis reveals substantial variations in the distribution of redundancy across diffusion steps among different DiT models. Interestingly, within a single model, the redundancy distribution remains stable regardless of variations in input prompts, step counts, or scheduling strategies.
Given the lack of a consistent pattern across diverse models, caching strategies designed for a specific group of models may not easily transfer to others. To overcome this challenge, we introduce a tool for analyzing the redundancy of individual models, enabling subsequent research to develop tailored caching strategies for specific model architectures. The project is publicly available at \url{https://github.com/xdit-project/DiTCacheAnalysis}

\end{abstract}


\section{Introduction}
\label{sec:introduction}

In recent years, diffusion models have emerged as predominant tools in the realms of image and video generation owing to their profound comprehension of user prompts, capacity for high-resolution generation, and alignment with real-world consistency. A notable trend in model architecture involves a transition from the conventional U-Net backbone~\cite{ronneberger2015u} to the innovative Diffusion Transformer (DiT)~\cite{peebles2023scalable}. Concurrently, the parameter count in DiT models grows remarkably, enabling the handling of higher-resolution images and videos with increased frame counts. However, the expansion in both model parameters and input sequence length has significantly escalated the inference latency of DiT models. The generation of a single high-resolution image typically demands minutes due to this heightened latency, thereby bringing not only challenges for model deployment and application but also unfeasibility to real-time services.

In response to this challenge, numerous research is focused on mitigating the inference latency of DiT models. Leveraging the transformer backbone of DiT, existing methodologies aimed at accelerating transformer computations, such as pruning~\cite{castells2024ld}, distillation~\cite{geng2024one}, quantization~\cite{li2023q}, and parallelization~\cite{fang2024pipefusionpatchlevelpipelineparallelism, fang2024xditinferenceenginediffusion} can be repurposed. These efforts aim to enhance time efficiency for each diffusion step, thereby reducing the overall latency. Furthermore, a separate line of research has identified similarities in activations between adjacent steps in the diffusion process, identifying a concept known as redundancy~\cite{ma2024learning, zhang2024cross, zhao2024real}. Researchers in this domain have proposed adopting a caching mechanism, wherein model outputs from early diffusion steps are preserved and later reused to minimize computational overhead. By leveraging these methodologies, researchers aim to streamline the inference process of DiT models, paving the way for more efficient and practical applications in real-world scenarios.

While diffusion transformers (DiT) have demonstrated promising results, there remains a notable gap in exploring the impact of the DiT structure on generation, alongside the absence of a tailored acceleration framework for the DiT architecture. Some existing literature has analyzed a limited set of DiT models, derived observations, and subsequently proposed caching methods to align with their findings. Despite achieving speedups in the models under study, the observations from prior research lack consistency, with some findings even conflicting with each other. This variability hampers the scalability of methods developed for one model to be applied effectively to others. This inconsistency not only perplexes researchers in the field, hindering the development of new methodologies, but also prompts a reassessment of the value of studying redundancy given the divergent outcomes from previous research endeavors.

In our study, we start by reviewing existing efforts in redundancy evaluation before embarking on a systematic exploration of redundancy across diffusion steps. We consider a broad range of DiT models, input prompts, the number of diffusion steps, and scheduler selections. Our findings reveal that redundancy manifests varied distributions across different models. However, within a single model, the redundancy distribution remains consistent and is stable to variations in user prompts, inference steps, and schedulers. Consequently, we conclude that previous observations derived from a limited model set cannot be universally applied. Rather than pursuing a generalized caching method for accelerating DiT inference, we suggest designing cache mechanisms tailored to each individual model.

To facilitate this approach, we have also released a lightweight tool enabling future researchers to swiftly examine the redundancy distribution specific to the model they are investigating, thereby facilitating the seamless design of caching strategies to accelerate DiT inference processes.

In summary, we make the following contributions in this paper:

\begin{itemize}
    \item We conduct a thorough investigation into the redundancy discussed in the existing literature.
    \item We systematically explore redundancy across diffusion steps in a diverse range of prominent DiT models, including FLUX.1-dev, Pixart-Alpha, Stable-Diffusion-3, CogVideoX-5B, Open-Sora, Latte-1, and Mochi-1-preview.
    \item We rigorously examine the redundancy distribution under various prompts, differing numbers of diffusion steps, and various schedulers, and then present our empirical results.
    \item We introduce a software tool \textit{DiTCacheAnalysis} designed to aid future researchers in developing model-specific caching strategies.
\end{itemize}

\section{Background \& Related Work}
\label{sec:background}

\textbf{Diffusion Models}: Diffusion models utilize a noise-prediction deep neural network (DNN) denoted by $\epsilon_{\theta}$ to generate a high-quality image.
The process starts from pure Gaussian noise $x_T \sim \mathcal{N}(0, I)$ and involves numerous iterative denoising steps to produce the final meaningful image $x_0$, with $T$ representing the total number of diffusion time steps. 
At each diffusion time step $t$, given the noisy image $x_t$, the model $\epsilon_{\theta}$ takes $x_t$, $t$, and an additional condition $c$ (e.g., text, image) as inputs to predict the corresponding noise $\epsilon_t$ within $x_t$. 
At each denoising step, the previous image $x_{t-1}$ can be obtained from the following equation:
\begin{equation}
    x_{t-1} = \text{Update}(x_t, t, \epsilon_t), \quad \epsilon_t = \epsilon_{\theta}(x_t, t, c).
\end{equation}

In this context, \text{Update} denotes a function that is specific to the sampler, i.e. DDIM~\cite{song2020denoising} and DPM~\cite{lu2022dpm}, generally involves operations such as element-wise operations.
After multiple steps, we decode $x_0$ from the Latent Space to the Pixel Space using a Variational Autoencoder (VAE)~\cite{kingma2013auto}.
Consequently, the predominant contributor to diffusion model inference latency is attributed to the forward propagation through the model $\epsilon_{\theta}$.

\textbf{Diffusion Transformers (DiTs)}: The architecture of diffusion model $\epsilon_{\theta}$ is undergoing a pivotal transition from U-Net~\cite{ronneberger2015u} to Diffusion Transformers (DiTs)~\cite{jiang2024megascale, chen2023pixart, ma2024latte}, driven by the scaling law demonstrating increased model parameters and training data with enhanced model performance. 
Unlike U-Nets, which apply convolutional layers capturing spatial hierarchies, DiTs segment the input into latent patches and leverage the transformer's self-attention mechanism to model relationships within and across these patches.

In DiTs, the input noisy latent representation is decomposed into patches. These patches are embedded into tokens and fed into a series of DiT blocks.
DiT blocks generally incorporate Multi-Head Self-Attention, Layer Norm, and Pointwise Feedforward Networks.
However, the specifics of different DiT models may vary. 
For instance, the incorporation of conditioning can be achieved through various methods such as adaptive layer norm~\cite{peebles2023scalable}, cross-attention~\cite{chen2023pixart}, and extra input tokens~\cite{esser2024scaling}. 
As diffusion models tackle higher-resolution images and longer visual sequences, they impose a quadratic computational burden on inference.

\textbf{Input Temporal Redundancy:}
Diffusion Model entails the iterative prediction of noise from the input image or video. Recent research has highlighted the concept of \textit{input temporal redundancy}, which refers to the similarity observed in both the inputs and activations across successive diffusion timesteps~\cite{so2024frdiff, ma2024deepcache}. A recent study~\cite{ma2024learning} further investigates the distribution of this similarity across different layers and timesteps. Based on the redundancy, a branch of research caches the activation values and reuses them in the subsequent diffusion timesteps to prune computation. For example, in the context of the U-Net architecture, DeepCache updates the low-level features while reusing the high-level ones from cache~\cite{ma2024deepcache}. By contrast, in the domain of DiT models, L2C~\cite{ma2024learning}, \textsc{Tgate}~\cite{zhang2024cross}, and PAB~\cite{zhao2024real} explores the redundancy distribution across diffusion steps and propose caching mechanism to accelerate the diffusion process. We will describe the findings in these three methods in detail in Section \ref{sec:existing}. Besides, $\Delta$-DiT proposes to cache the rear DiT blocks in the early sampling stages and the front DiT blocks in the later stages~\cite{chen2024delta}. DiTFastAttn~\cite{yuan2024ditfastattn} identifies three types of redundancies, such as spatial redundancy, temporal redundancy, and conditional redundancy, and presents an attention compression method to speed up generation. Finally, PipeFusion employs a sequence-level pipeline parallel strategy to manage communication and computation processes efficiently by leveraging input temporal redundancy~\cite{fang2024pipefusionpatchlevelpipelineparallelism, fang2024xditinferenceenginediffusion}.



\section{Previous Observations on DiT's Input Temporal Redundancy}
\label{sec:existing}

This section provides an overview of three seminal works that investigate the input temporal redundancy within DiTs.

\textbf{L2C.~\cite{ma2024learning}} 
To the best of our knowledge, L2C is the first method studying the input temporal redundancy between consecutive diffusion steps in DiT inference. L2C compares the difference in activation values of each DiT layer between diffusion steps, and finds that the performance varies significantly across steps, even at the same layer. The difference is markedly higher in the later steps compared to the early denoising steps, and is greater in multi-head attention layers than in feed-forward layers.

\textbf{\textsc{Tgate}.~\cite{zhang2024cross}} \textsc{Tgate} endeavors to unveil the functionality of cross-attention in the diffusion process of a pre-trained DiT. By examines the disparities in cross-attention maps in SD-2.1 between two consecutive diffusion steps, \textsc{Tgate} unveils a distinct trend characterized by elevated values in the initial stages that gradually converge towards zero within the 5 to 10 step range.

\textbf{PAB.~\cite{zhao2024real}} Addressing the substantial computational overhead inherent in video generation, PAB investigates the attention discrepancies between diffusion steps in video generation models like Open-Sora. The findings reveal a discernible U-shaped pattern, with values approximating zero in intermediate steps while peaking at both ends. Various attention mechanisms, such as spatial, temporal, and cross-attention, exhibit diverse degrees of stability throughout the diffusion process.

\begin{figure}[htbp!]
    \centering
    \begin{subfigure}[t]{0.32\textwidth}
        \centering
        \includegraphics[height=1.2in]{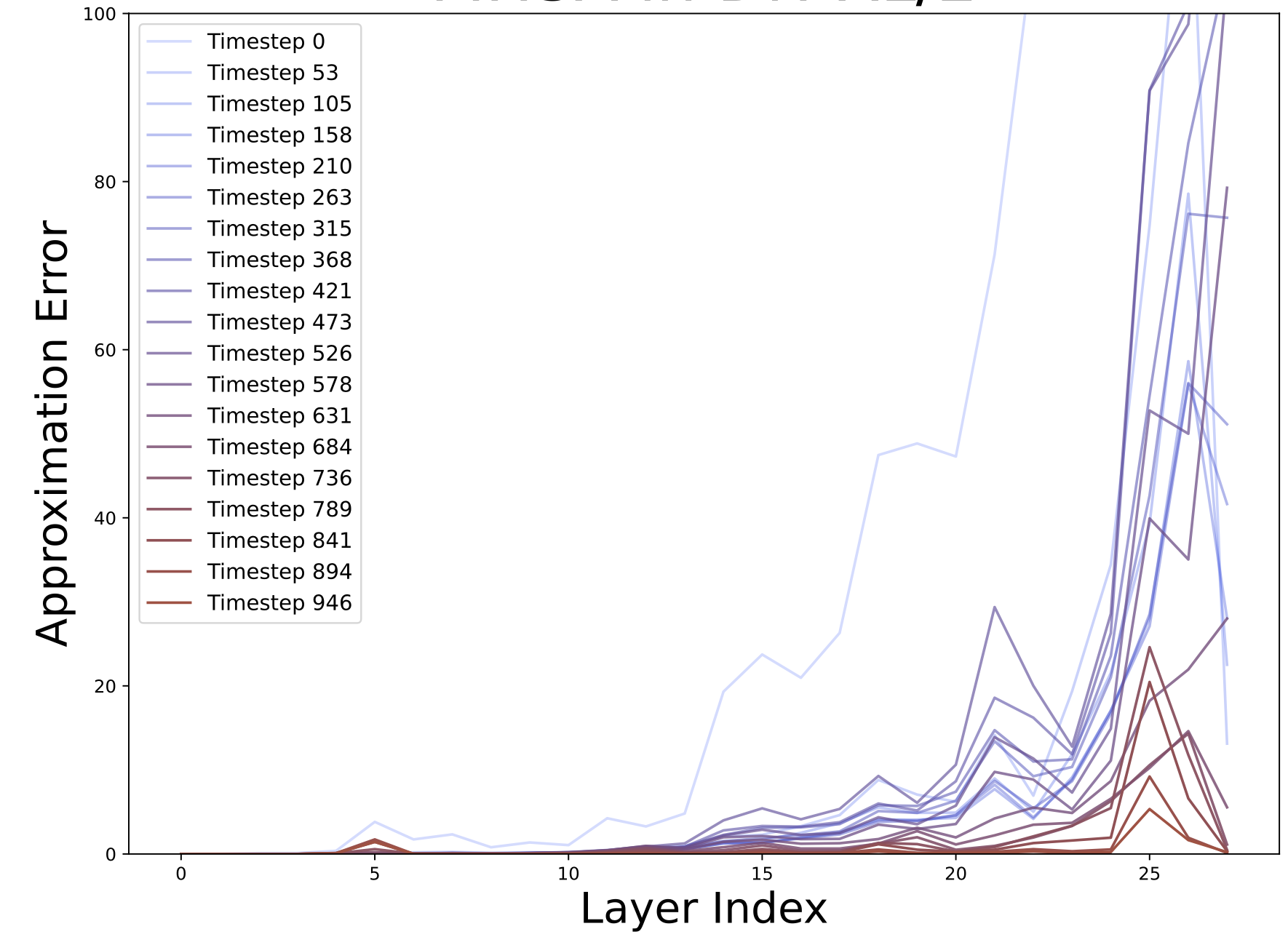}
        \caption{L2C.}
    \end{subfigure}
    \begin{subfigure}[t]{0.32\textwidth}
        \centering
        \includegraphics[height=1.2in]{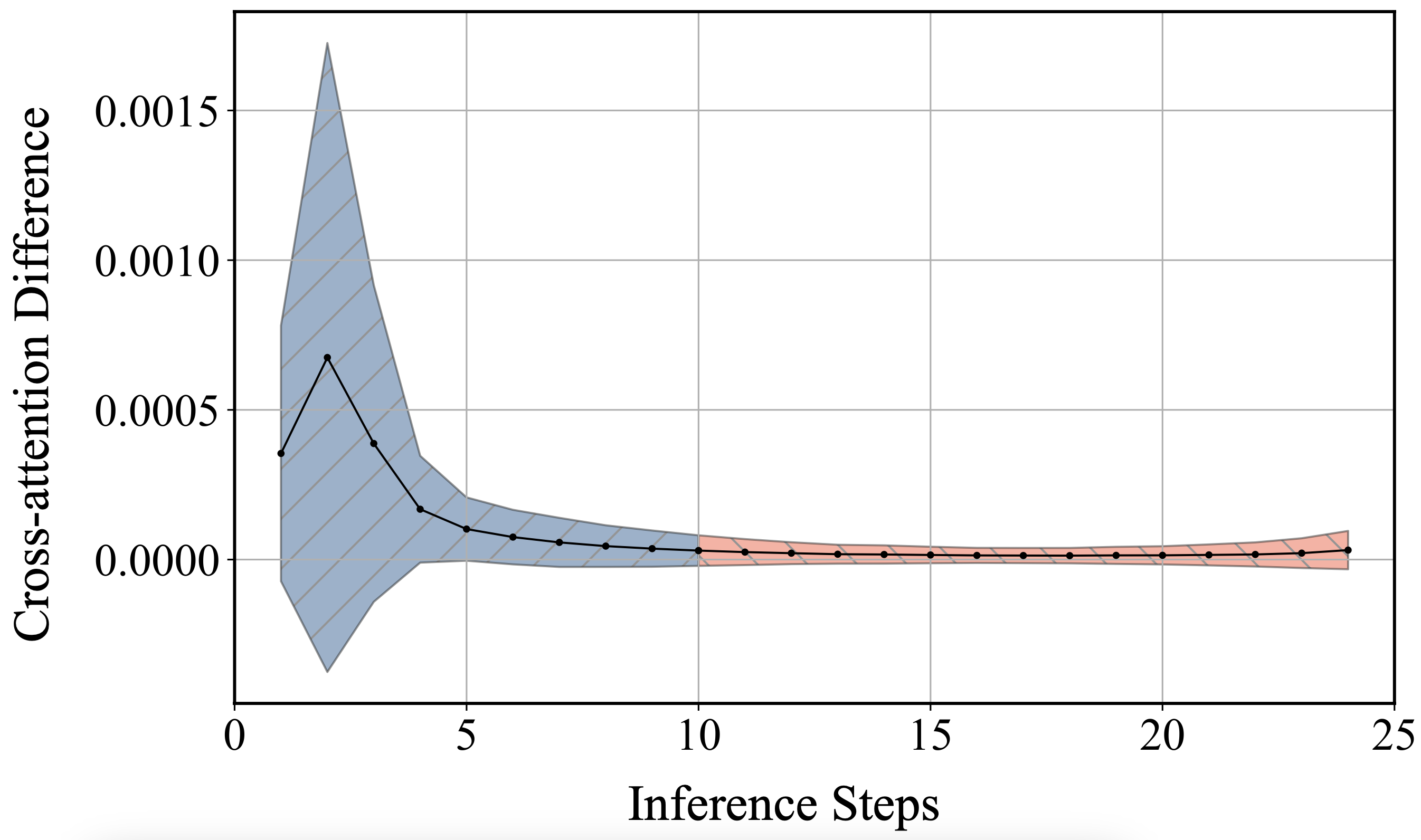}
        \caption{\textsc{Tgate}.}
    \end{subfigure}
    \begin{subfigure}[t]{0.32\textwidth}
        \centering
        \includegraphics[height=1.2in]{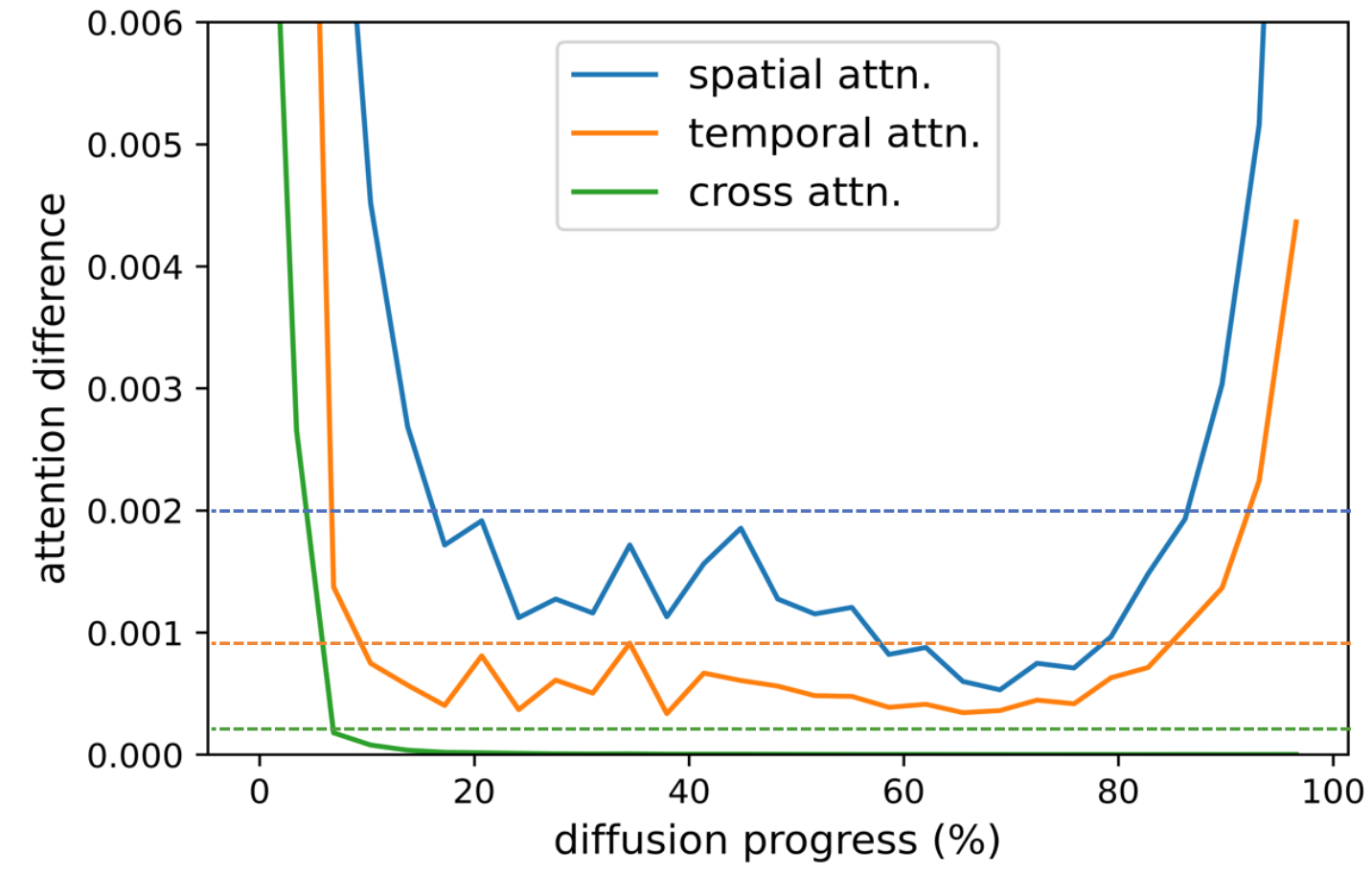}
        \caption{PAB.}
    \end{subfigure}
    \caption{Figures in previous research.}
    \label{fig:previous}
\end{figure}

Figure \ref{fig:previous} shows the results from these papers.

\section{Our Systematic Study}

As evident from Section \ref{sec:existing}, disparities exist in previous observations regarding the redundancy distribution among diffusion steps, with certain conclusions even conflicting. Consequently, this section undertakes extensive experiments focusing on input temporal redundancy.

\subsection{Experimental Setup.}

To make our work systematic, we incorporate a diverse array of mainstream DiT models, each introduced below:

\begin{itemize}
    \item \textbf{Flux.1-dev.~\cite{flux}} Flux.1-dev, a rectified flow transformer with 12 billion parameter, excels in generating images from textual descriptions.. 
    \item \textbf{Pixart-Alpha.~\cite{pixart}} Pixart-Alpha comprises pure transformer blocks tailored for latent diffusion, enabling the direct generation of 1024px images from textual prompts within a single sampling process.
    \item \textbf{Stable-Diffusion-3.~\cite{sd3}} Stable-Diffusion-3, a Multimodal Diffusion Transformer (MMDiT) text-to-image model, boasts enhanced performance in image quality, typography, prompt comprehension, and resource efficiency.
    \item \textbf{CogVideoX-5B.~\cite{cogvideo}} CogVideoX serves as an open-source video generation model.
    \item \textbf{Open-Sora.~\cite{opensora}} Open-Sora represents a dedicated model initiative aimed at efficiently producing high-quality videos.
    \item \textbf{Latte-1.~\cite{latte}} Latte-1, a latent diffusion transformer for video generation.
    \item \textbf{Mochi-1-preview.~\cite{mochi}} Mochi-1 preview, an open-state-of-the-art video generation model, exhibits high-fidelity motion and robust adherence to prompts in preliminary evaluations.
\end{itemize}

The number of parameters, number of attention layers, and default number of diffusion steps are listed in Table \ref{tab:0-setup}

\begin{table}[htbp!]\footnotesize
 \caption{Coefficient of variation in L1 distance of $K$, $V$, and $A$ over various prompts.}
  \centering
  \resizebox{\textwidth}{!}{%
  \begin{tabular}{lccccccc}
    \toprule
    Metrics & Flux.1-dev & Pixart-Alpha & Stable-Diffusion-3 & CogVideoX-5B & Open-Sora & Latte-1& Mochi-1-preview\\
    \midrule
    \#Parameters       & 12B & 0.6B & 2B & 5B & 1.1B & 0.7B & 10B \\
    \#Attention Layers & 57  &  28  & 24 & 30 & 56 & 56 & 48 \\
    \#Diffusion Steps  & 28  &  20  & 28 & 50 & 30 & 50 & 64 \\
    \bottomrule
  \end{tabular}
  }
  \label{tab:0-setup}
\end{table}

For our experiments, we utilize the captions of the initial 1000 images from the MS-COCO dataset as prompts for generating images or videos with these models.For each attention layer of each model, we record the input $K$ and $V$ for attention computation, along with the activation output $A$ of the attention layer. We denote $X^{i,j}_{\mathcal{M}(p)}$ ($X \in \{K, V, A\}$) as the value of $X$ in Layer $j$ at the $i$th diffusion step when the model $\mathcal{M}$ is generating images or videos for prompt $p$. Our study on redundancy focuses on the difference of $K$, $V$, and $A$ between consecutive diffusion steps. Therefore, we further denote $\Delta X^{i,j}_{\mathcal{M}(p)}$ ($X \in \{K, V, A\}$) as the L1 distance between $X^{i,j}_{\mathcal{M}(p)}$ and $X^{i+1,j}_{\mathcal{M}(p)}$. Note that, in the diffusion process, Step $i+1$ appear before Step $i$. And a lower L1 distance indicates a high degree of redundancy.

\subsection{Varying Prompts}

Given model $\mathcal{M}$, layer $j$, and diffusion step $i$, we calculate a list of L1 distance for $X$ ($X \in \{K, V, A\}$), denoted as $[\Delta X^{i,j}_{\mathcal{M}(p_1)}, \Delta X^{i,j}_{\mathcal{M}(p_2)}, \cdots, \Delta X^{i,j}_{\mathcal{M}(p_N)}]$, where each element corresponds to a prompt in $[p_1, p_2, \cdots, p_N]$. To investigate whether the input prompt influences the redundancy distribution, we compute the coefficient of variation over the array of L1 distance. This coefficient is defined as the ratio of the standard deviation to the mean of the array. Subsequently, we calculate the average coefficient of variation across all layers and diffusion steps for each model, presenting the results in Table \ref{tab:1-varying-prompts}. The table reveals that the choice of prompt impacts the L1 distance of $K$, $V$, and $A$ by approximately 10\% to 15\% on average. This suggests that while the DiT model predominantly determines the redundancy distribution, the contribution from prompts is limited. Therefore, in subsequent experiments, instead of examining redundancy for each prompt individually, we report the average L1 distance across all prompts to provide a comprehensive overview.

\begin{table}[htbp!]\scriptsize
 \caption{Coefficient of variation in L1 distance of $K$, $V$, and $A$ over various prompts.}
  \centering
  \resizebox{\textwidth}{!}{%
  \begin{tabular}{lccccccc}
    \toprule
    Target & Flux.1-dev & Pixart-Alpha & Stable-Diffusion-3 & CogVideoX-5B & Open-Sora & Latte-1 & Mochi-1-preview\\
    \midrule
    $K$ & 10.0\% & 10.4\% & 10.5\% & 13.9\% & 11.6\% & 12.6\% & \\
    $V$ & 10.0\% & 11.4\% & 11.1\% & 14.1\% & 13.7\% & 12.7\% & \\
    $A$ & 10.8\% & 12.7\% & 11.3\% & 14.9\% & 15.0\% & 14.1\% & \\
    \bottomrule
  \end{tabular}
  }
  \label{tab:1-varying-prompts}
\end{table}

\subsection{Varying DiT models}

This subsection compares the distribution of redundancy across diffusion steps in various DiT models.

\begin{figure}[htbp!]
\centering
\includegraphics[width=\textwidth]{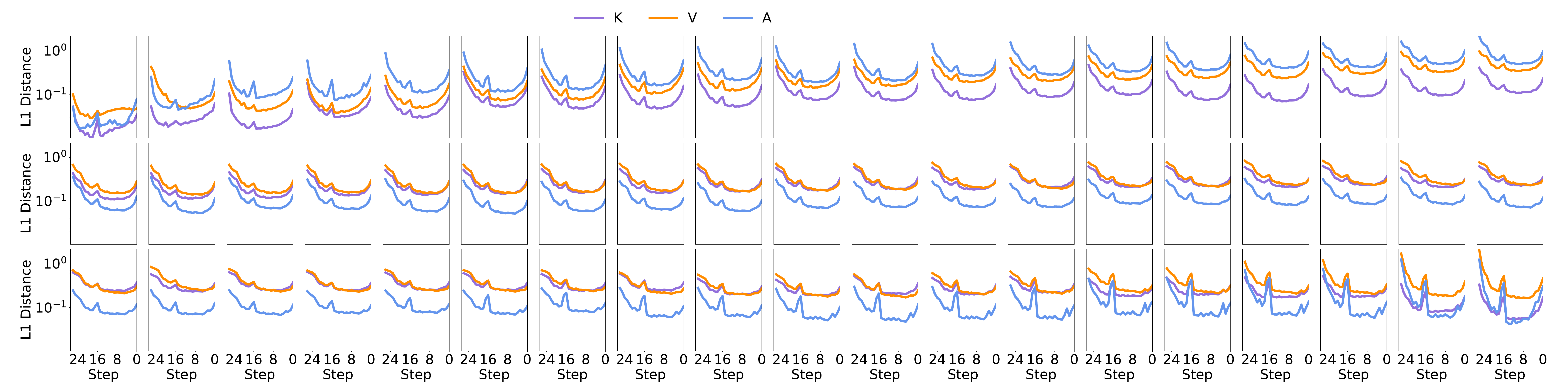}
\caption{Distribution of redundancy in FLUX.1-dev.}
\label{fig:2-flux}
\end{figure}

\textbf{Flux.1-dev.} Figure \ref{fig:2-flux} illustrates the distribution of L1 distance in $K$, $V$, and $A$ in Flux.1-dev. Each subfigure represents the L1 distance distribution over diffusion steps for a specific layer of the model. Overall, the redundancy in $K$, $V$, and $A$ follows a similar trend across layers, with higher values in early diffusion steps that decrease with progression. Notably, a significant change occurs around the 16th diffusion step, leading to a sharp rise in L1 distances for $K$, $V$, and $A$. Additionally, differences in L1 distances are more pronounced in deeper layers compared to shallow layers.

\begin{figure}[htbp!]
\centering
\includegraphics[width=\textwidth]{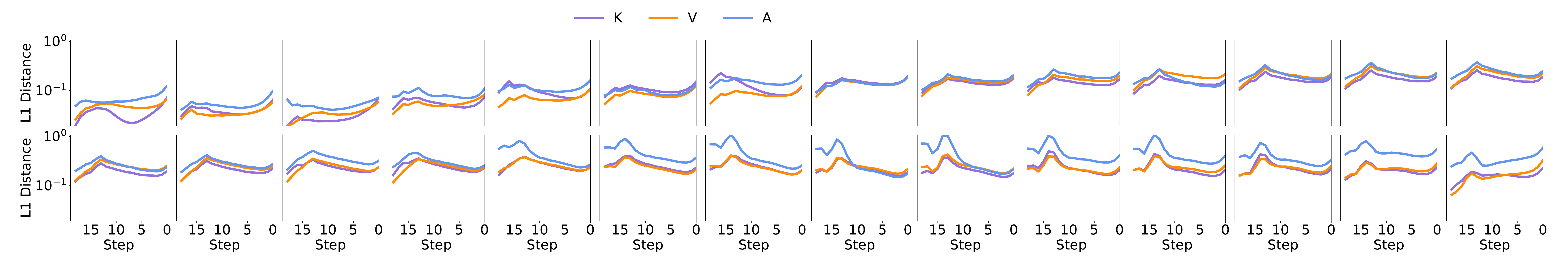}
\caption{Distribution of redundancy in Pixart-Alpha.}
\label{fig:2-pixart}
\end{figure}

\textbf{Pixart-Alpha.} Figure \ref{fig:2-pixart} showcases the results for Pixart-Alpha, exhibiting a distinct trend from FLUX.1-dev. In shallow layers, the L1 distances of $K$, $V$, and $A$ are very close to zero. However, in deep layers, these distances start small, increase notably from the 16th to 10th step, and then decrease towards the end of the diffusion process. Moreover, the L1 distance differences for $A$ are significantly higher than those for $K$ and $V$.

\begin{figure}[htbp!]
\centering
\includegraphics[width=\textwidth]{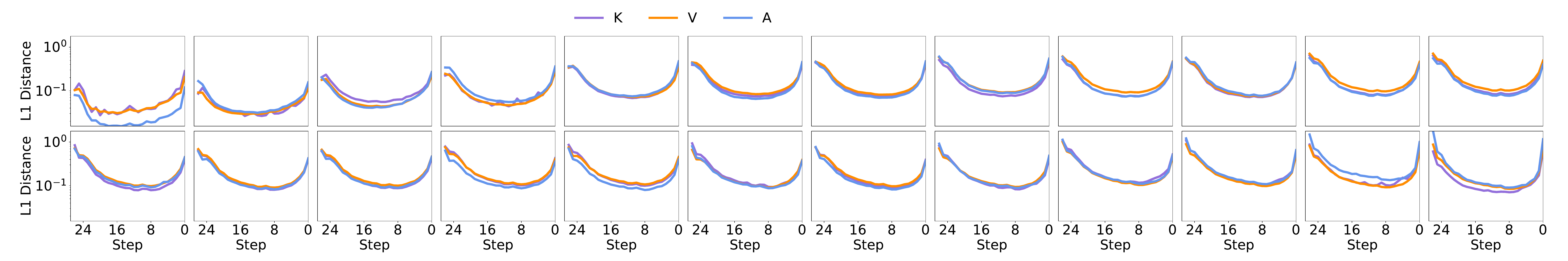}
\caption{Distribution of redundancy in Stable-Diffusion-3.}
\label{fig:2-sd3}
\end{figure}

\textbf{Stable-Diffusion-3.} As depicted in Figure \ref{fig:2-sd3}, the L1 distance differences in L1 distance in Stable-Diffusion-3 exhibit a U-shaped pattern, with higher differences at the beginning and end of each subfigure and lower differences in the middle diffusion steps. Notably, L1 distances for $K$, $V$, and $A$ show similar patterns, with differences escalating in deeper layers of the model.

\begin{figure}[htbp!]
\centering
\includegraphics[width=\textwidth]{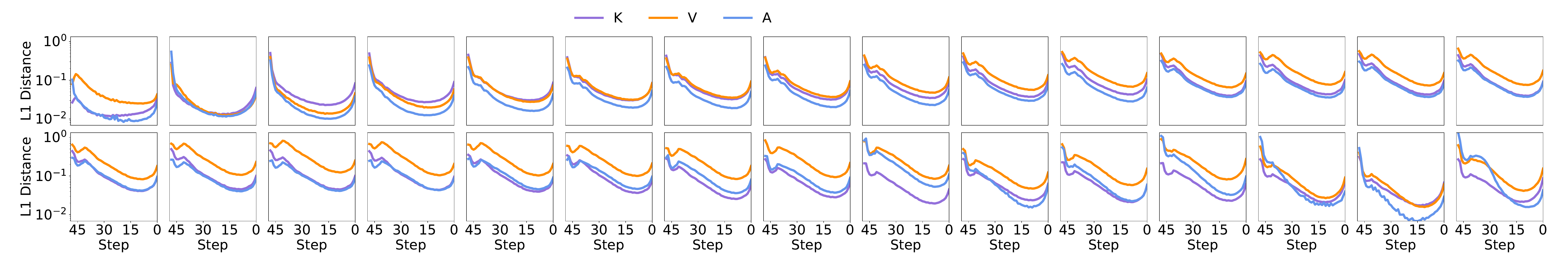}
\caption{Distribution of redundancy in CogVideoX-5B.}
\label{fig:2-cogvideo}
\end{figure}

\textbf{CogVideoX-5B.} Figure \ref{fig:2-cogvideo} displays the L1 distance distribution in $K$, $V$, and $A$ for CogVideoX-5B. The distances initially decrease in the first 5 steps, then increase between steps 5-10, and finally decrease towards the end of the diffusion process. Unlike Pixart-Alpha, $K$ and $A$ exhibit more redundancy compared to $V$.

\begin{figure}[htbp!]
\centering
\includegraphics[width=\textwidth]{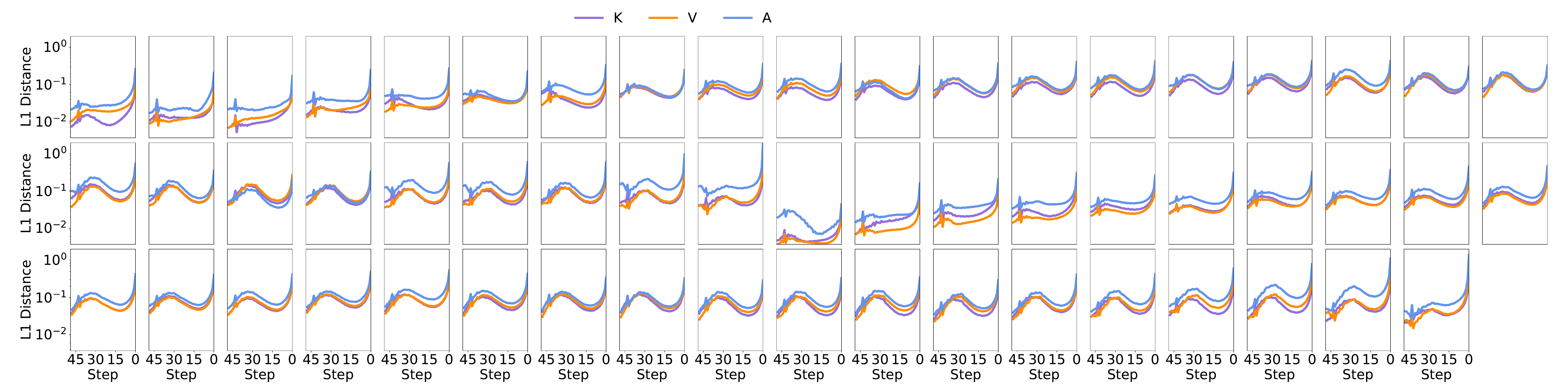}
\caption{Distribution of redundancy in Latte-1.}
\label{fig:2-latte}
\end{figure}

\textbf{Latte-1.} In Figure \ref{fig:2-latte}, the Latte-1 model demonstrates a high level of redundancy across most diffusion steps, except for a significant increase in L1 distances at several steps towards the end.

\begin{figure}[htbp!]
\centering
\includegraphics[width=\textwidth]{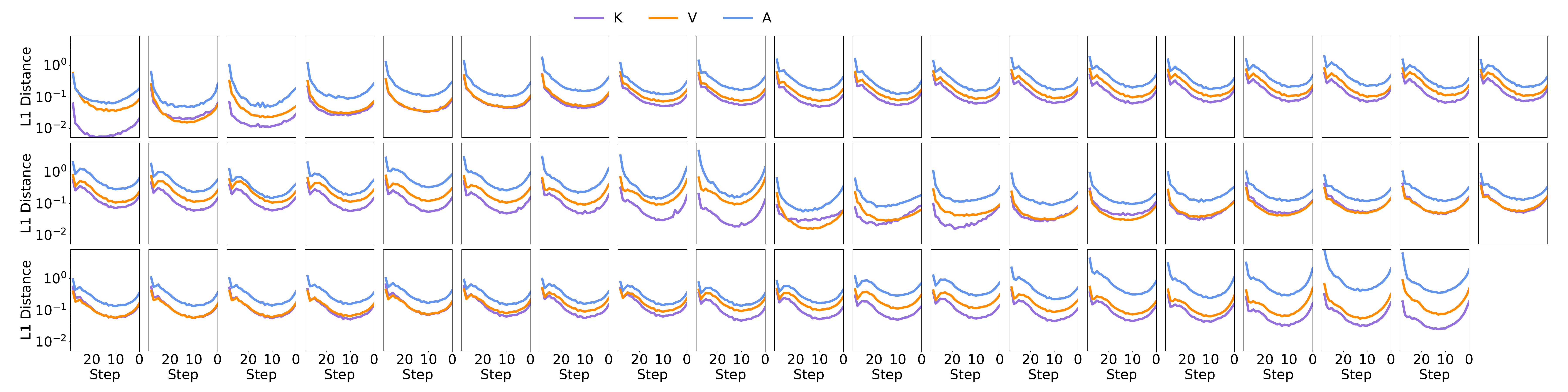}
\caption{Distribution of redundancy in Open-Sora.}
\label{fig:2-opensora}
\end{figure}

\textbf{Open-Sora.} Figure \ref{fig:2-opensora} reveals the L1 distance distribution in $K$, $V$, and $A$ for Open-Sora. Notably, the L1 distances in $A$ are considerably higher than those in $K$ and $V$ throughout the diffusion process. For $K$ and $V$, distances are higher in the initial steps compared to the later steps. In contrast, L1 distances in $A$ exhibit a substantial increase at the beginning and end of the diffusion process, significantly exceeding those in the middle steps.

\begin{figure}[htbp!]
\centering
\includegraphics[width=\textwidth]{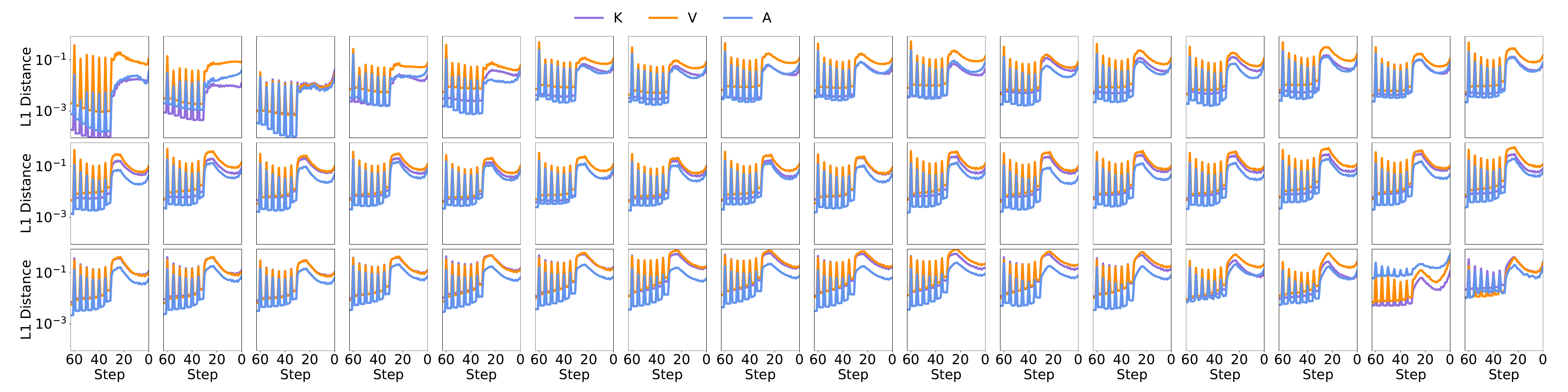}
\caption{Distribution of redundancy in Mochi-1-preview.}
\label{fig:2-mochi}
\end{figure}

\textbf{Mochi-1-preview.} As the inference latency of Mochi-1-preview is high, we utilize the first 100 captions from the 1000 selected ones as prompts. The redundancy distribution is depicted in Figure \ref{fig:2-mochi}. A significant fluctuation is evident in the initial stages of the diffusion process, where the L1 distances of $K$, $V$, and $A$ exhibit abrupt shifts between 0 and values ranging from 0.1 to 0.5, varying across different layers. Conversely, in the latter part of the diffusion process, the disparities in $K$, $V$, and $A$ initially peak before gradually diminishing.

These results highlight the substantial variability in redundancy distribution among different DiT models, indicating that no single observation from prior studies can encompass all DiT models. To ascertain whether redundancy trends are model-dependent or influenced by diffusion configurations, we conduct further ablation studies as outlined below.

\subsection{Varying Number of Diffusion Step}

\begin{figure}[htbp!]
\centering
\includegraphics[width=\textwidth]{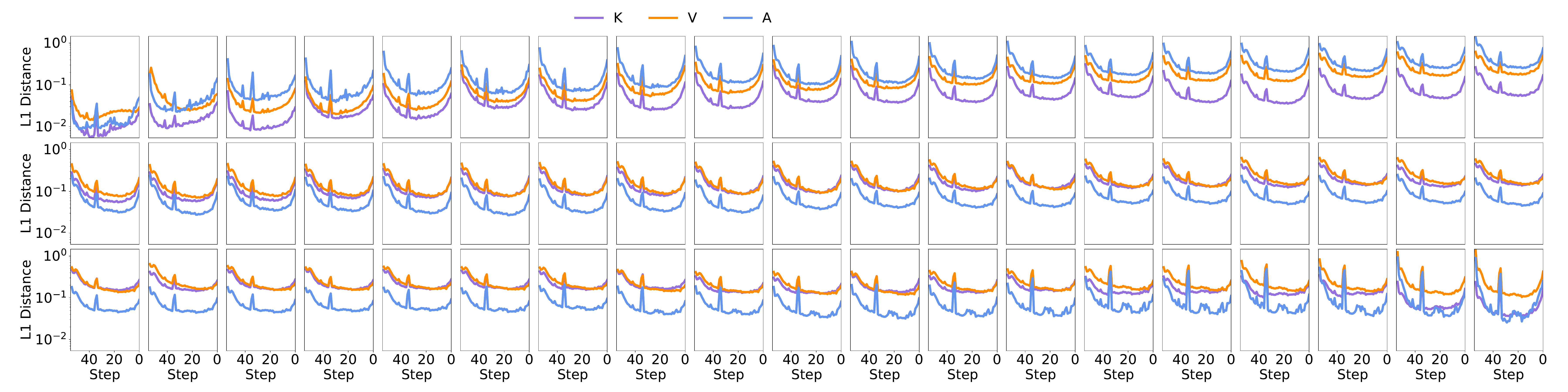}
\caption{Distribution of redundancy in FLUX.1-dev with 56 diffusion steps.}
\label{fig:3-flux-56}
\end{figure}

\begin{figure}[htbp!]
\centering
\includegraphics[width=\textwidth]{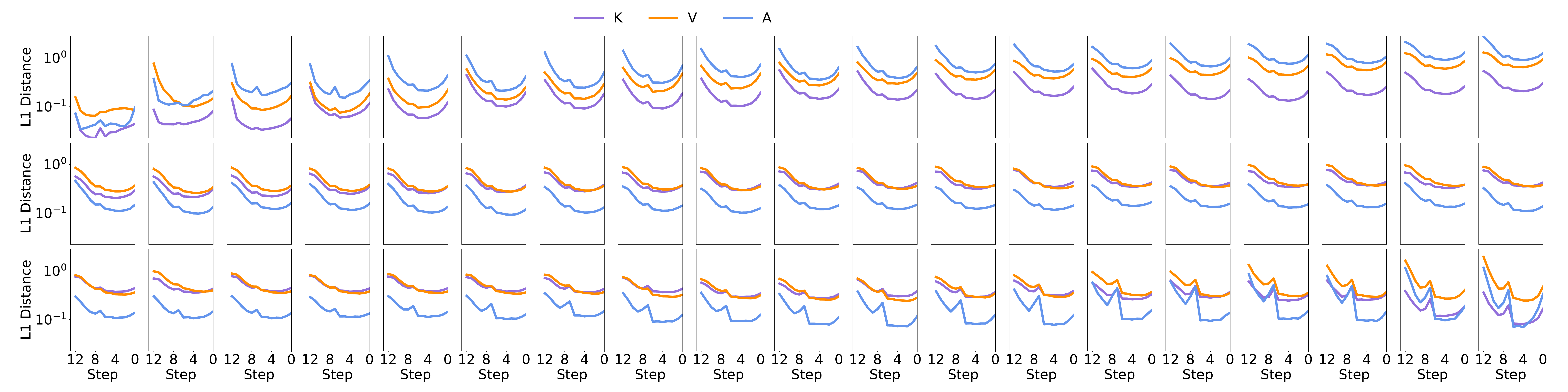}
\caption{Distribution of redundancy in FLUX.1-dev with 14 diffusion steps.}
\label{fig:3-flux-14}
\end{figure}

In the FLUX.1-dev model, the default number of inference step is 28. To investigate the impact of the length of the diffusion process on redundancy distribution, we conducted two additional experiments on FLUX.1-dev by varying the number of inference steps to 56 and 14 while keeping all other settings constant. Figure \ref{fig:3-flux-56} and \ref{fig:3-flux-14} depict the trends of L1 distance across all layers in the model, respectively. 

Analysis of the figures reveals that when using 56 diffusion steps, the overall trends closely resemble those observed in the default setting. Notably, a significant change in L1 distances occurs around the 34th diffusion step, maintaining a consistent relative position throughout the diffusion process.

On the other hand, when reducing the number of inference steps to 14, the plots in the figures exhibit smoother patterns compared to those with 28 or 56 steps. However, the trends in L1 distances remain stable. Consequently, we can infer that the number of diffusion steps does not significantly alter the redundancy distribution in DiT models.

\subsection{Varying Scheduler}

\begin{figure}[htbp!]
\centering
\includegraphics[width=\textwidth]{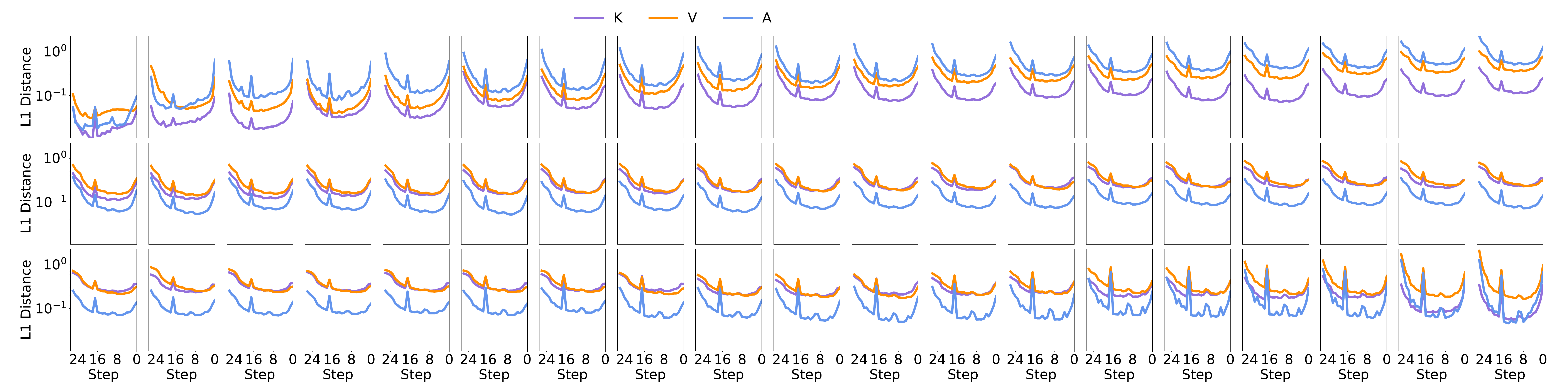}
\caption{Distribution of redundancy in FLUX.1-dev with Stable-Diffusion-3's scheduler.}
\label{fig:4-flux-sd3-scheduler}
\end{figure}

Upon observing that both FLUX.1-dev and Stable-Diffusion-3 employ the Flow Match Euler Discrete Scheduler with distinct configurations, we opted to replace the scheduler of FLUX.1-dev with that of Stable-Diffusion-3 to explore its impact on redundancy in diffusion. Figure \ref{fig:4-flux-sd3-scheduler} closely resembles Figure \ref{fig:2-flux} with the default scheduler, indicating minimal differences.

Consequently, our investigation suggests that the choice of scheduler does not have a notable influence on the distribution of redundancy in DiT models.

\subsection{Conclusion}

This study delves into the examination of redundancy within Diffusion-in-Transformer (DiT) models throughout the diffusion process. Commencing with a thorough review of existing research insights, we progress to present our systematic investigation. Encompassing a diverse array of mainstream DiT models, we conduct ablation studies involving varying prompts, models, diffusion steps, and schedulers. The key findings derived from our experimental analyses are outlined as follows:
\begin{enumerate}
    \item The redundancy distribution within each DiT model exhibits distinct trends.
    \item In the context of a singular model, the redundancy distribution demonstrates stability irrespective of alterations in input prompts, step counts, or scheduling strategies.
\end{enumerate}

Our results suggest that caching strategies tailored for specific subsets of models may not seamlessly transfer to others. Consequently, we introduce a lightweight tool for exploring redundancy in DiT models, facilitating the development of customized caching strategies tailored to specific model architectures in subsequent research endeavors.

\textbf{Future work}: This paper aims to clarify the redundancy distribution to aid researchers in diminishing the latency of individual diffusion steps through appropriate approximations. However, it would be interesting to investigate whether we can maintain comparable generation quality while decreasing the number of inference steps without resorting to approximations. Furthermore, rather than crafting caching methods tailored to each distinct DiT model based on its redundancy patterns, an engaging avenue for research involves dynamically capturing redundancy behavior during DiT operation and adjusting the caching strategy based on ongoing observations.

\bibliographystyle{unsrt}  
\bibliography{references}

\end{document}